\documentclass[letterpaper, 10pt, conference]{ieeeconf}      

\usepackage[pdftex]{graphicx}
\usepackage{float}
\floatstyle{plaintop}
\restylefloat{table}
\usepackage[tableposition=top]{caption}
\restylefloat{table}
\usepackage{textcomp}
   \DeclareGraphicsExtensions{.pdf,.jpeg,.png}

\IEEEoverridecommandlockouts                              

\usepackage{amsmath} 

\title{\LARGE \bf
A Robust Lane Detection and Departure Warning System
}

\author{Mrinal Haloi$^{1}$ and Dinesh Babu Jayagopi$^{2}$
\thanks{$^{1}$Mrinal Haloi is with Dept. Electronics and Communication Enginering, IIT Guwahati
        {\tt\small h.mrinal@iitg.ernet.in or mrinal.haloi11@gmail.com};  Copyright\textcopyright 2015 by the authors}%
\thanks{$^{2}$Dinesh Babu Jayagopi is with IIIT Bangalore
        {\tt\small jdinesh@iiitb.ac.in}}%
}

\begin{document}

\maketitle
\thispagestyle{empty}
\pagestyle{empty}

\begin{abstract}

In this work, we have developed a robust lane detection 
and departure warning technique. 
Our system is based on single camera sensor. For lane detection a modified 
Inverse Perspective Mapping using only a few extrinsic camera parameters and 
illuminant Invariant techniques is used. Lane markings are represented using 
a combination of 2nd and 4th order steerable filters, robust to shadowing. 
Effect of shadowing and extra sun light are removed using Lab color space, 
and illuminant invariant representation. Lanes are assumed to be cubic curves 
and fitted using robust RANSAC. This method can reliably detect lanes of the 
road and its boundary. This method has been experimented in Indian road conditions under different challenging situations
and the result obtained were very good. For lane departure angle an optical flow based 
method were used.
\end{abstract}

\section{INTRODUCTION}
With the increasing number of vital life loses in accidents, 
India is one of the most accident prone country, where according 
to the NCRB report 135,000 died in 2013 and property damage of 
\$ 20 billion \cite{news}. Many a time, accidents and unusual traffic 
congestion take place due to careless and impatient nature of drivers. 
In most cases drivers don't follow lane rules, traffic rules leading to 
traffic congestion and accidents. For counter measuring all these problems, 
advanced driver assistance system is needed that can assist people drive 
safely or drive itself safely in case of autonomous driving cars. It is quite
a challenge to make autonomous car that can self-sense the environment and 
drive like an aware human. In some recent works researchers have developed 
autonomous car, even though it’s not still deployable in real life.

\begin{figure}[h!]
  \centering
      \includegraphics[width=3.3in,height=2.4in]{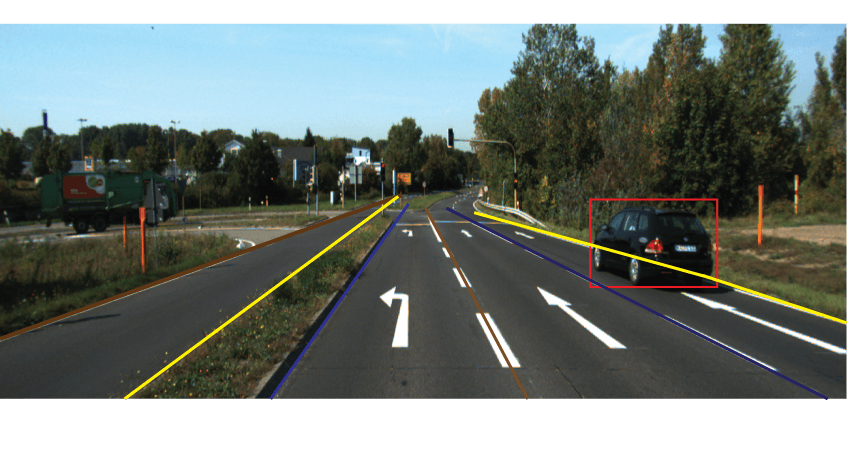}
\caption{Vehicle lane detection scenarios}
\label{fig:intro}
\end{figure}
In developed countries such as U.S.A and Germany, with the gradual emergence of 
autonomous driving research, efforts are on to build a smart driving system 
that can drive more safely without any fatigue, as compared to humans can be 
programmed to follow traffic rules. In this context the main challenge involves 
understanding complex traffic patterns and taking real time decision on the basis of visual 
data from camera and laser sensors. For these automatic cars, modelling 
vehicle current lane with respect to the road environment is very relevant for 
accurate driving, maintaining correct lane and keeping track of front vehicles. 
Apart from this estimating departure angle from the current lane for possible 
overtaking and taking turn in curved road is also important. Fig.~\ref{fig:intro} shows a sample
vehicle lane detection scenario. 

Whether automatic cars will become a future reality or not, Advanced Driver Assistance
Systems (ADAS) are increasingly deployed in modern cars. Road lane, road boundary
and departure angle estimation are crucial modules of ADAS. Such modules facilitate 
and validate human judgment. They can also be used to prepare the automobile in case
of an emergency situation. 

In this work, we have developed a robust lane detection system using a modified 
Inverse Perspective Mapping (IPM) algorithm where camera intrinsic parameters are not required. This 
IPM formulation can give up to 45m of accurate road view. By using IPM a wide area of unwanted region 
is removed and this will help to locate lane features accurately. Also a novel lane 
departure warning system is developed. In addition to that another contribution of 
our method is that it can detect road boundary lane even if there are no lane marking, 
suitable for developing countries context. For this purpose wide angle 
camera sensor mounted on car roof was used to capture surrounding road environment. 

Rest of the paper is organised as follows, Section II describes related literature.
In Section III elaborate detection of lane and computation of departure angle method, and in Section IV
our experimental setup. Finally we conclude in Section V. 

\section{Related Work}
The related literature on autonomous driving and advanced driving assistance system 
are based on using single or multiple camera facing 
the road to detect lane feature. Sensors specifically LIDAR, RADAR etc. are used for detecting 
object and 3D modelling of the road environment. Also in some works driver behaviour 
understanding using a camera facing the driver is used for drowsiness, sleepiness, and 
fatigue detection. 

For lane detection and departure warning system many works 
target urban and countryside area \cite{c2},\cite{c3}, \cite{c4}, \cite{c5}, \cite{c6}, \cite{c7}. 
Most of these works have used Image Processing and Machine Learning based 
approach for defining and extracting features for lanes. Some of the works 
have straight and planer road assumption and only works in highway. This happens
because of assumption of strong lane marking and low traffics. 
In lane tracking kalman filter and particle filter [8] based approach is used. 
Another work based on vanishing point detection [9] gives a good result in countryside road. 

Other related works include methods for road segmentation, traffic signs 
detection and recognition, 3D modelling of road environment (e.g. \cite{c10},\cite{c13},\cite{c14}). 
Parallax flow computation was used by Baehring et al. for detecting overtaking and close 
cutting vehicles \cite{c11}. For detecting and avoiding Collison, Hong et al. had used Radar, 
LIDAR, camera and omnidirectional camera respectively \cite{c12}, \cite{c16}. They focused on 
detecting using LIDAR sensor data classifying object as static and dynamic and tracking 
using extended Kalman filter and for getting a wide view of surrounding situation. 
For detection of forward collision Srinivasa et al. have used forward looking camera and radar data \cite{c13}. 

In some works driver behaviour and inattentiveness was modelled using fatigue detection, 
drowsiness, eye tracking, visual analysis etc. Ji et al. \cite{c17} presented tracking method for eye, 
gaze and face pose and Hu et al. \cite{c18} used SVM based method for driver drowsiness detection. Driver behavior was modelled using visual analysis of surrounding environment in our previous work \cite{c15}.

\section{Our approach}

\begin{figure}[h!]
  \centering
      \includegraphics[width=3.5in,height=2.7in]{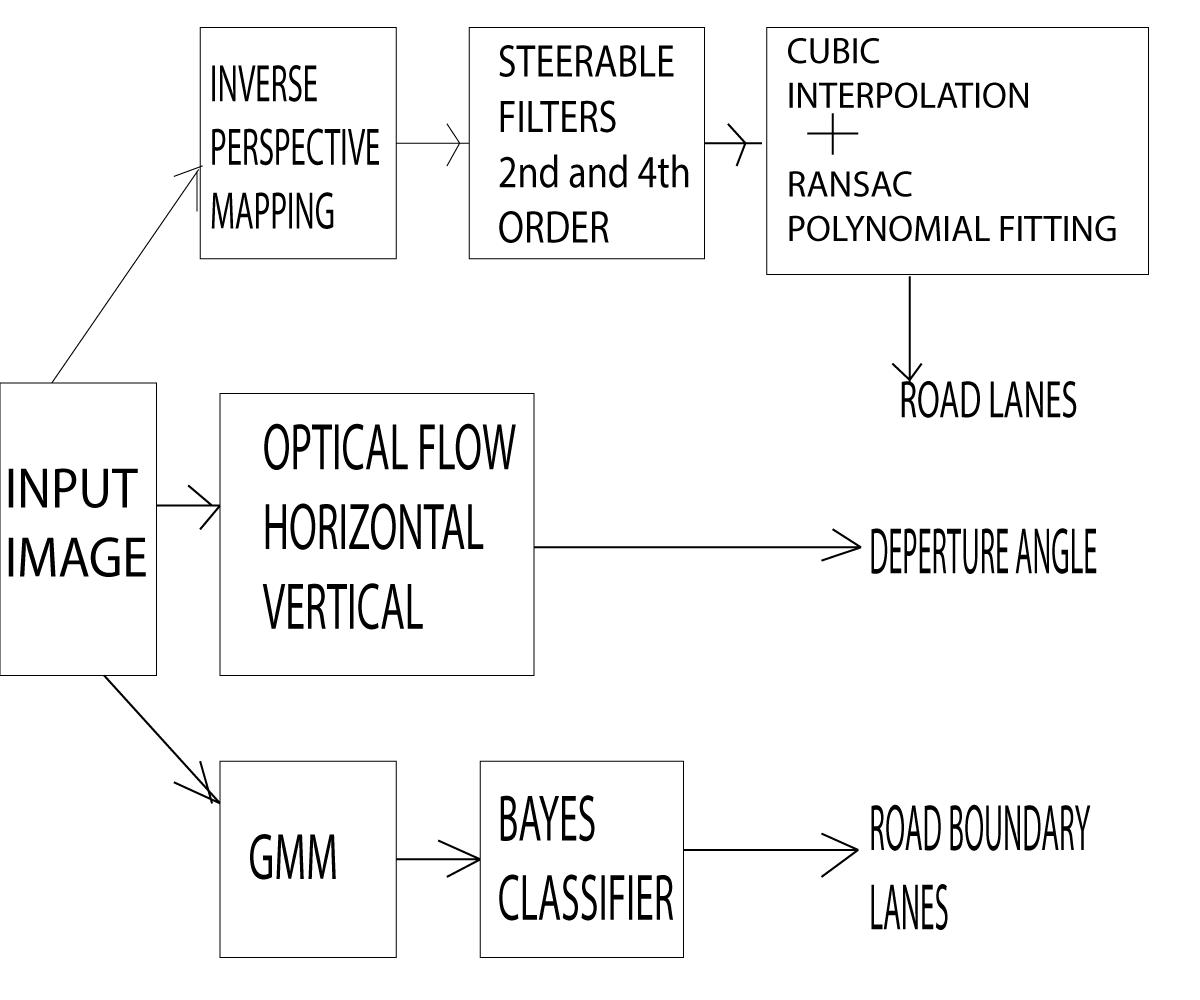}
\caption{Overview of Method}
\label{fig:blck}
\end{figure}

We summarize our system in Fig.~\ref{fig:blck}, as to how we estimate road lanes, road boundary and 
departure angle. 

\subsection{Lane detection}
For localising vehicle on the road estimation of some related parameters like its current lane, 
shape of the road and its position from centreline. To compute this parameters the road 
environment is scanned using wide angle camera sensor and extract lane markers. For lane detection we have 
proposed a novel method using Lab colour space, 2nd and 4th order steerable filters and improved 
Inverse Perspective Mapping. Below our lane markers extraction algorithm is described.

\begin{figure}[h!]
  \centering
      \includegraphics[width=3.5in,height=3.1in]{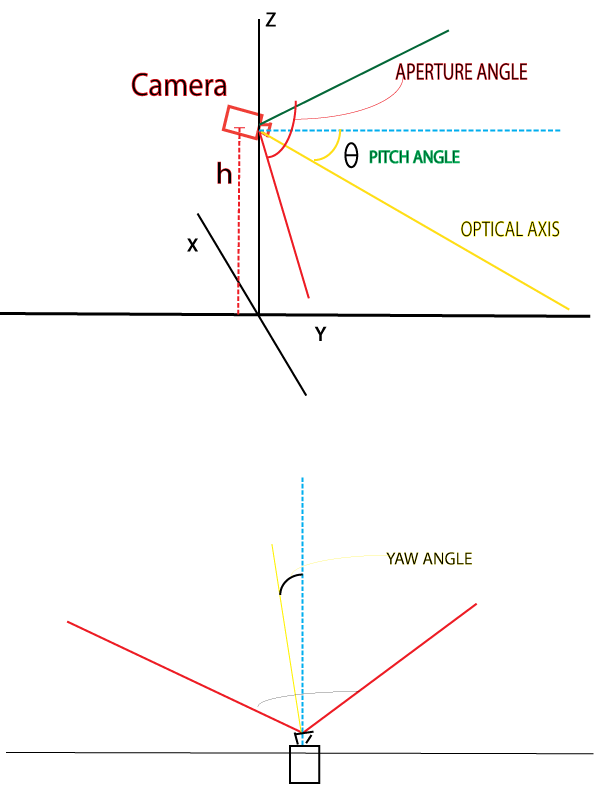}
\caption{Camera Setup, pitch angle, Yaw angle}
\label{fig:cam}
\end{figure}

{A1. Perspective effect} \\
In real world situation if two parallel lines are captured they appears to be converged to some distant points so their nature can't be understood in images. Road lanes are parallel, for detecting and localising them in images the effect of perspective projection is removed using Inverse Perspective Projection. In this work we have presented a modified version of IPM,  which is robust to a distance of 45m. No internal parameter calibration of camera is required for computation in comparison to other algorithm \cite{c2}, \cite{c14}. Suppose Camera location with respect to car coordinates system (C{x},C{y},C{z}) where C{z} will be the height from ground lane 'h'. Optical axis make an angle $\theta$ known as pitch angle, $\gamma$ yaw angle and $\alpha$ as half of camera aperture as shown in Fig.~\ref{fig:cam}.

\begin{figure}[h!]
  \centering
      \includegraphics[width=3.5in,height=1.7in]{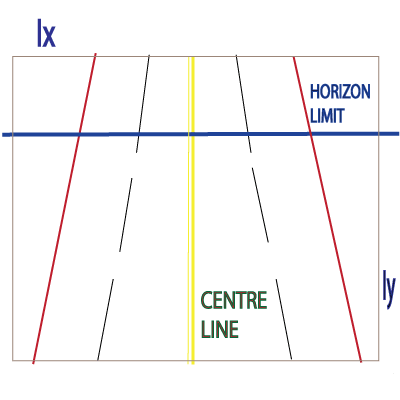}
\caption{Area of interest}
\label{fig:horz}
\end{figure}

To increase computational speed removing uninterested area 
from image we define horizon line from where our interested area will lie below "Horizon Limit" as shown in Fig.~\ref{fig:horz}. After applying this horizon limit, inverse perspective mapping will be applied to this modified image.

 For derivation of IPM road is assumed to be perfectly planer. 
 Because of this assumption different type of obstacles present in the 
 road which does not lie in the road are deformed and seen as some noisy 
 area in IPM image. Denote image coordinates as $(Ix,Iy)$ and real world coordinates as $(x,y,0)$. 
 Suppose camera resolution as $m \times n$, then the following image to world frame mapping equation is obtained from derivation
\begin{equation}
\delta = tan^{-(m-1)/sqrt((m-1)^2 + (n-1)^2)*tan(\alpha)}
\end{equation}
\begin{equation}
\omega = tan^{-(n-1)/sqrt((m-1)^2 + (n-1)^2)*tan(\alpha)}
\end{equation}
\begin{equation}
hz = \frac{(m-1)*0.5}{(1-tan(\theta)/tan(\delta))} + 1
\end{equation}
\begin{equation}
x = h\frac{1+(1-2\frac{Ix-1}{m-1})tan(\delta)tan(\theta)}{tan\theta-(1-2\frac{Ix-1}{m-1})tan(\delta)}
\end{equation}

\begin{equation}
y = h\frac{(1-2\frac{Iy-1}{n-1})tan(\omega)}{sin\theta-(1-2\frac{Ix-1}{m-1})tan(\delta)cos\theta}
\end{equation}

where $hz$ represent start row of image of interest.

\begin{figure}[h!]
  \centering
      \includegraphics[width=3.5in,height=2.35in]{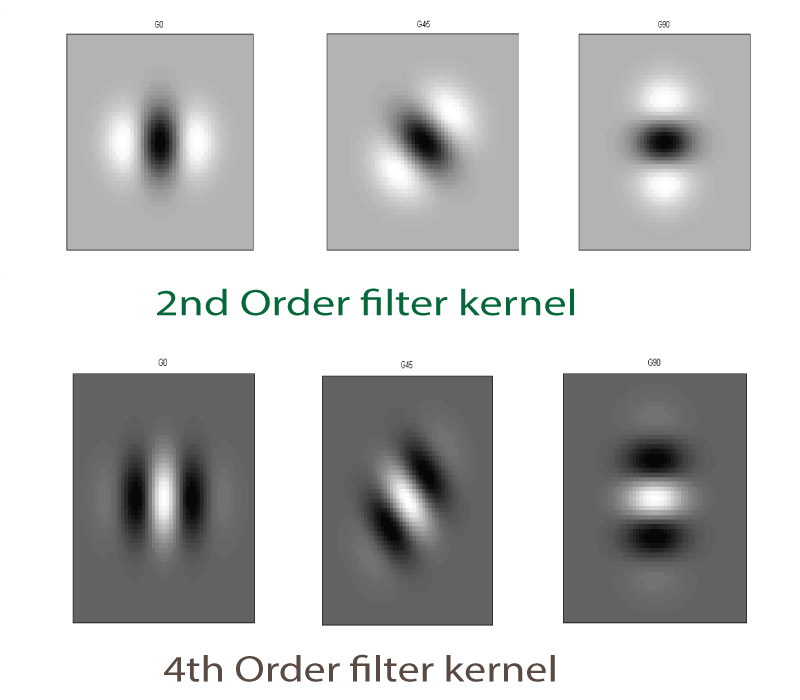}
\caption{Response at 0,45 and 90 degree}
\label{fig:fil}
\end{figure}

{A2. Feature Extraction} \\
For detecting lines and curves, 2D steerable filters \cite{c19} are very 
effective to use, because gradient changes due to color variation of 
road and lanes is effectively captured. In addition to that, due to their 
separability nature, computation is faster than other filters. 
The result obtained from both 2nd and 4th order filters are combined to extract final lane markings on the basis of adaptive 
thresholding, this depends also values of gradient angle to supress 
edge in unwanted direction. In Fig.~\ref{fig:fil} filters used in this method is shown.
Filter kernel used are represented by Eq. (6) to (10).

\begin{equation}
G_{2x}(x,y) = (\frac{4x^{2}}{\sigma^{4}} - \frac{2}{\sigma^{2}})e^{-\frac{x^{2} + y^{2}}{\sigma^{2}}}
\end{equation}
\begin{equation}
G_{xy}(x,y) = \frac{4xy}{\sigma^{4}}e^{-\frac{x^{2} + y^{2}}{\sigma^{2}}}
\end{equation}
\begin{equation}
G_{2y}(x,y) = (\frac{4y^{2}}{\sigma^{4}} - \frac{2}{\sigma^{2}})e^{-\frac{x^{2} + y^{2}}{\sigma^{2}}}
\end{equation}

\begin{equation}
G_{4x}(x,y) = (\frac{16x^{4}}{\sigma^{8}} - \frac{48x^{2}}{\sigma^{6}} - \frac{12}{\sigma^{4}})e^{-\frac{x^{2} + y^{2}}{\sigma^{2}}}
\end{equation}
\begin{equation}
G_{4y}(x,y) = (\frac{16y^{4}}{\sigma^{8}} - \frac{48y^{2}}{\sigma^{6}} - \frac{12}{\sigma^{4}})e^{-\frac{x^{2} + y^{2}}{\sigma^{2}}}
\end{equation}

{A3. Cubic Interpolation and RANSAC} \\
RANSAC method was used for identifying and getting potential lane points position from extracted features points for fitting a parabolic curves. RANSAC is an iterative algorithm, it removes outliers and fit defined model to data point. Maximum of 8 curves (lane lines) can be identified by using this proposed method. 
In most of the roads except centre lane other lines are discontinuous, to get continuous edge to fit a polynomial in those plain areas cubic interpolation are very efficient. Cubic interpolation used in these setup depends on gradient value and direction. 
Our road model is given in Eq. (11), where $y_{0}$ is offset from vertical coordinate system and $a$,$b$,$c$ are lane parameters
.

\begin{equation}
Y = y_{0} + aX + bX^{2} + cX^{3}.
\end{equation}

\subsection{Road Boundary Lane}
Only lane line extraction can't give an overall idea about 
car position if road boundaries are not known. Most of the time road 
lane boundary are not so clear and even not paved mainly in Indian situation, 
to cope up with this we need to segment the road area. For this 3 
class based Gaussian mixture model was used for segmentation of the road region. Since IPM image's majority pixels are road part and cars and other obstacles present in the road area becomes noise in the IPM image, so GMM can be used efficiently for this task. Method is applied in illuminant invariant 45m accurate IPM image, this method perform efficiently for this purpose. 

Three clusters used for segmentation comprise of road region, surrounding natural scenes
and road obstacles. Predefined mean and covariance values for our clusters was used. 
For computation of these initial means and covariance, we collected separate patches from 
train images for these three categories and computed those values. This initialization 
gives us better result than random k means initialization.
An iterative expectation maximization based algorithm is used to compute final means, 
covariance and probability of each clusters in GMM.

At the end a Bayesian classification techniques Eq. (12) is used to classify each pixels in image.

\begin{equation}
p(x/c_{i}) = p(c_{i})*e^{-\frac{(x - m_{c_{i}})^2}{2\sigma_{c_{i}}^2}}
\end{equation}
Here $c_{i}$ denotes a class i, $m_{c_{i}}$ means and $\sigma_{c_{i}}$ variances and prior probability $p(c_{i})$ of the class.

Using vanishing point estimation rest of the road boundary beyond 45m of images, 
which is not covered in IPM image can be approximately modelled \cite{c9}. 

\subsection{Lane Departure Angle}
\begin{figure}[h!]
  \centering
      \includegraphics[width=3.5in,height=1.8in]{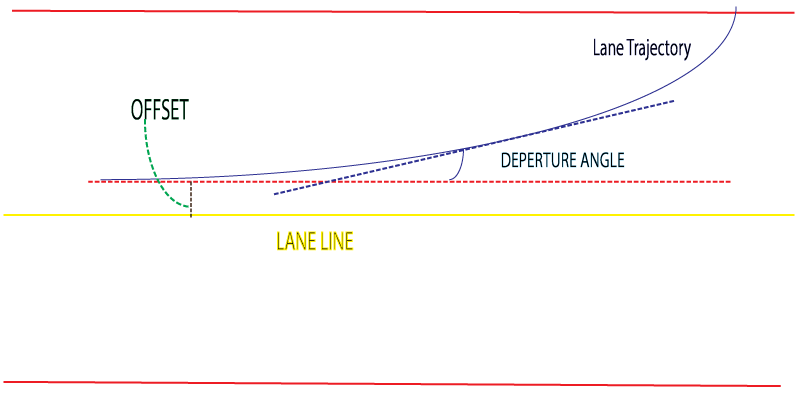}
\caption{Lane Deperture angle Idea }
\label{fig:traj}
\end{figure}

\begin{figure}[h!]
  \centering
      \includegraphics[width=3.5in,height=1.7in]{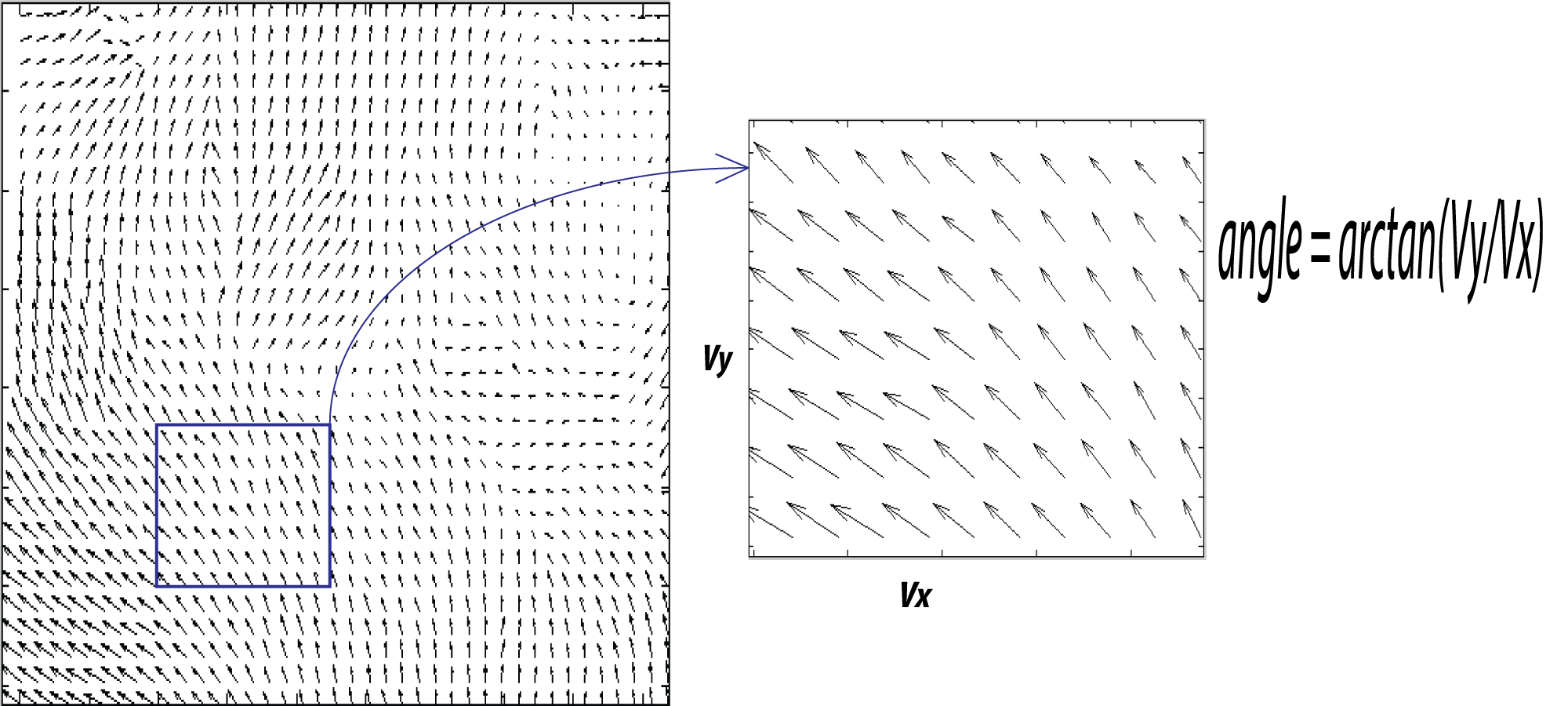}
\caption{Lane Deperture warning from optical flow}
\label{fig:optical}
\end{figure}

To avoid potential risk of accident or hazardous driving and maintaining proper lane, 
departure warning is very important. Using information from current position of 
vehicle with respect to lane specifically offset and optical flow computation 
this angle can be approximately computed \cite{c20}. Horizontal optical flow is a 
strong feature for knowing its unwanted horizontal velocity, which may be a
clear indication of lane changing or overtaking, except the case for curved road 
region. Fig.~\ref{fig:traj} and Fig.~\ref{fig:optical} illustrate the concept.
\begin{equation}
\Lambda = arctan(\frac{v_{y}}{v_{x}})
\end{equation}

\begin{figure*}

 \center

  \includegraphics[width=6.9in, height = 2.3in]{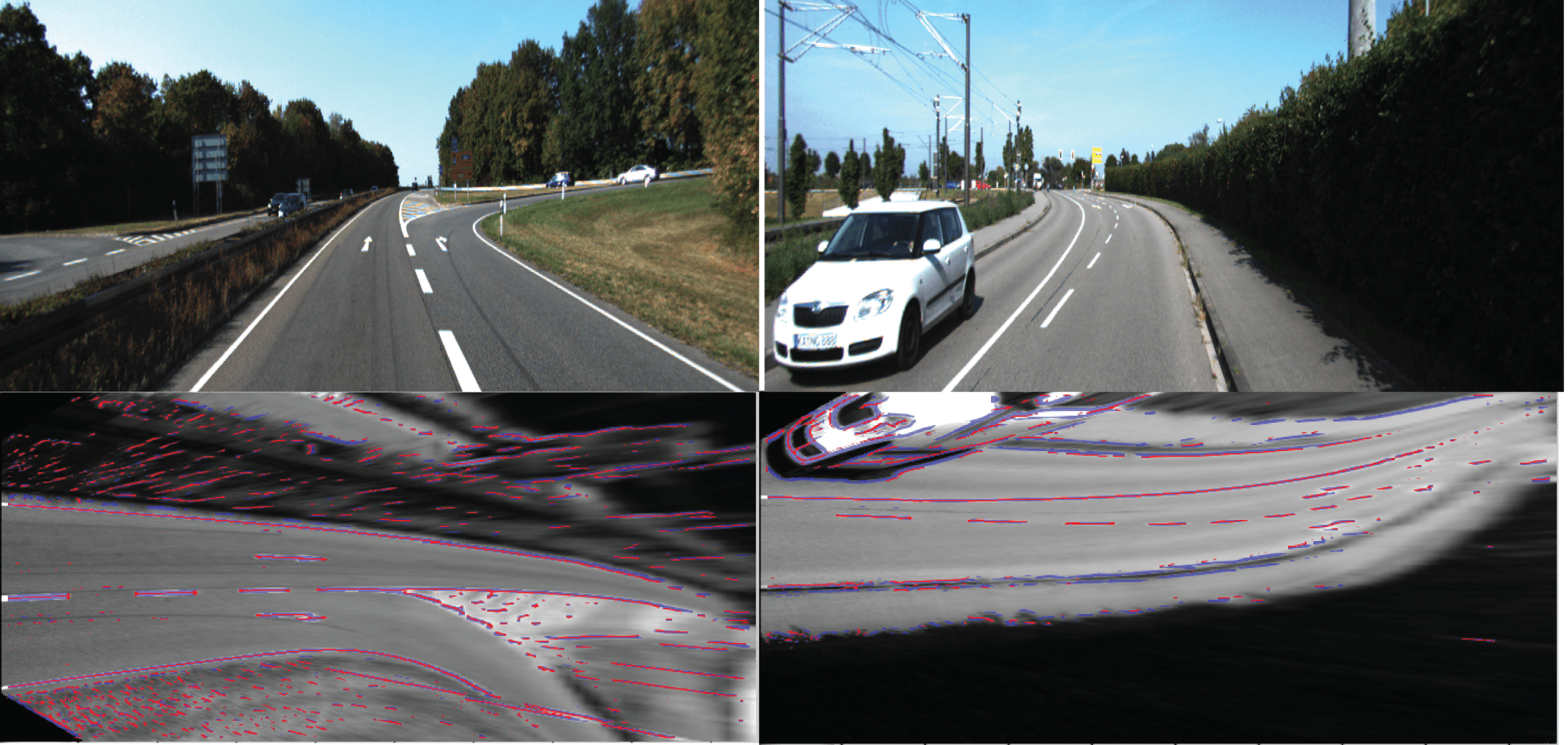}

  \caption{Image, IPM view and extracted lane features in KITTI dataset}

  \label{fig:kitti}

\end{figure*}
\begin{figure*}

 \center

  \includegraphics[width=6.9in,height=2.4in]{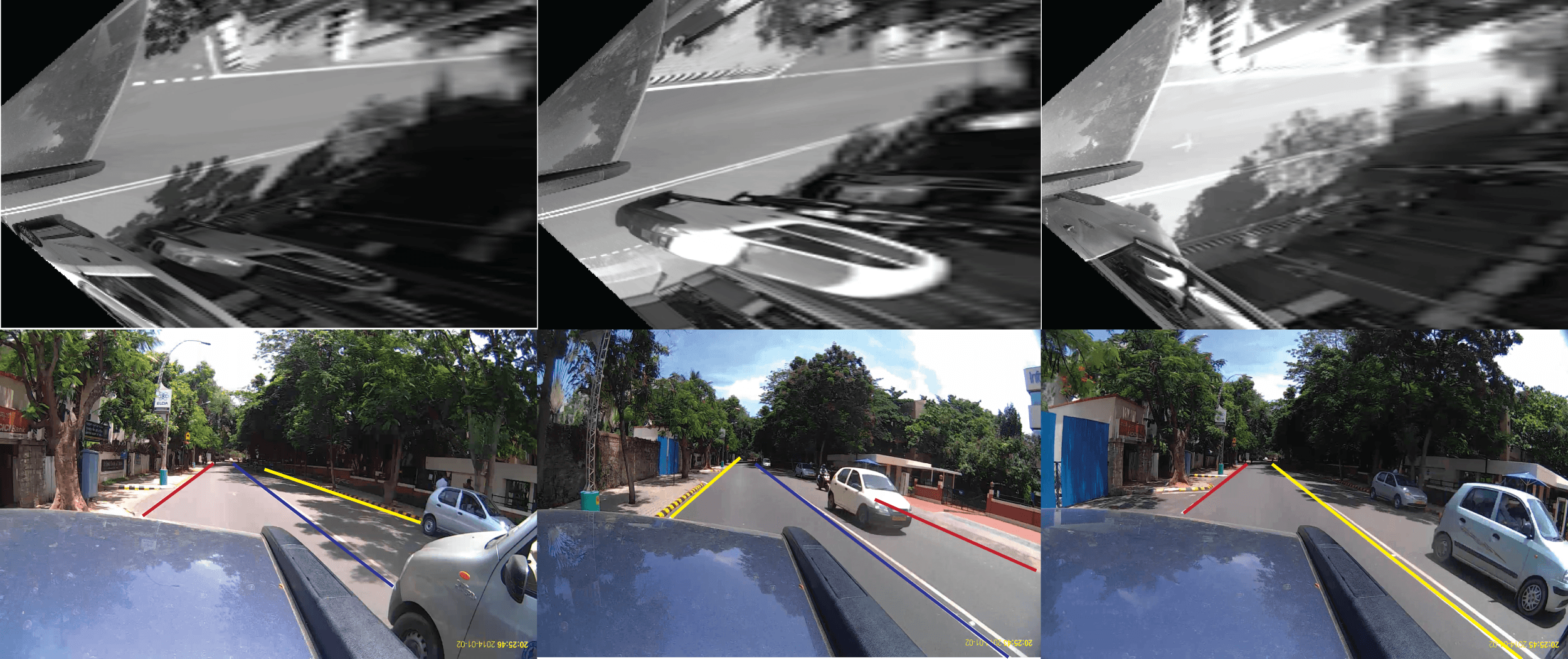}

  \caption{IPM image and detected lanes in indian road}

  \label{fig:india}

\end{figure*}

\section{EXPERIMENTAL RESULTS}
This proposed method has been experimented using different challenging datasets to analyze the performance and reliability.
Experimentation on image sizes of $440 \times 680$ and $375 \times 1242$  obtained from 
video of moving camera in different road environment were carried out. This system was developed using 
C++ language in LINUX intel quad core i7 machine.
We have collected a dataset in Bangalore city road condition, with a wide angle 
camera sensor mounted on our test vehicle's roof at height 155cm from ground 
plane at speed of around 45km/h pointing towards the forward road plane at an 
angle of $3\,^{\circ}{\rm }$ from horizontal line, for testing our algorithm 
accuracy in Indian condition. This dataset contains frames with varying luminance, shadows, curved lane lines and road without boundary lane lines. Also standard datasets KITTI \cite{c21}
and Caltech \cite{c14} was used for checking this algorithm in broad regions. These two datasets 
contain images with different condition like, sunny road with shadow, 
urban road with traffic and highway etc.

\begin{table*}[t]
  \centering
  \begin{tabular}{*{20}{c}}
\hline
{\bf Database} & {\bf \#Frame} & {\bf \#detectedAll} & {\bf \#Boundary}  &{\bf CorrectRate} & {\bf False Positive} & {\bf CorrectBoundary}\\
\hline
KITTI & 600 & 565 & 591 & 94.26 \% & 6.79 \% & 98.44 \%\\  
\hline
Caltech & 1224 & 1189 & 1204 & 97.14 \% & 4.17\% & 98.36 \% \\  
\hline
Indian Road & 1200 & 1087 & 1131 & 90.58 \% & 12.37\% & 94.25 \%\\
\hline
\end{tabular} 
  \caption{CorrectRate of ego-lane evaluation(upto 45m) and Road Boundary Detection}
\end{table*}

\begin{table*}[t]
  \centering
  \begin{tabular}{*{20}{c}}
 {\bf Method}  & {\bf PRE-20} & {\bf PRE-30}  & {\bf PRE-40}  & {\bf Runtime} & {\bf Environment}\\ 
\hline
 SPRAY [22] & 97.51\%  & 96.92 \%  & 88.76 \%  & 0.045 s & NVIDIA GTX 580 (Python + OpenCL)\\
\hline
Our Method & 95.17\% & 95.17\% & 93.76\% & 0.029s & 4 core @ 2.3 Ghz(C++)\\
\hline
BL [21] & 95.65 \% & 94.47 \%  & 87.23 \% & 0.02 s & 1 core @ 2.5 Ghz (Python)\\
\hline
SPlane + BL [23] & 95.48 \%  & 92.34 \%  & 79.79\%  & 2 s & 1 core @ 3.0 Ghz (C/C++)\\
\hline
\end{tabular} 
  \caption{Comparison with other Methods in KITTI dataset}
\end{table*}

With the combination of 2nd and 4th order steerable filters to detect edge 
in horizontal direction and vertical direction, results in the reduction of extra outliers, which help in robust 
fitting of lane lines and better input features to main RANSAC outliers removal and 
fitting lane lines. In addition to that horizontal lines on the road are associated with the detection of pedestrian crossing. This method is also capable of generating pedestrian crossing warning by detecting lines in horizontal direction. First row of Fig.~\ref{fig:kitti} shows the original images from KITTI dataset and second row depicts corresponding IPM image overlaid with extracted possible features points. In Fig.~\ref{fig:india} first row depicts the IPM images of indian road dataset and their corresponding final lane markings obtained after applying the algorithm is shown in second row. 
This method can detect road boundary even if there are no boundary lane marking, 
which is very useful for Indian road conditions. From Fig.~\ref{fig:kitti}, it can be seen 
that this method does not make any assumption of straight road, can detect 
lane features even if road is not straight. We have tested our algorithm in 
Indian road condition and obtained acceptable accuracy, also it can detect road boundary 
very well and the road region in road where there are no lane marking exist. 

We have observed that using illuminant invariance techniques, separating 
luminance and colour parts of image gives better accuracy over normal 
RGB images for better detection of lane lines. This setting are useful for various illumination changes due to shadows, raining, fogs etc.

In Table 1 we have given an analysis of precision (correct rate), false positive and correct boundary, obtained in lane detection and 
road boundary detection in three dataset. Pixel wise evaluation was used for the computation of these parameters. 
Lane detection and road boundary detection are building  blocks for better accuracy of this method. Also robustness due to lack of boundary lane and shadow can be observed in Fig.~\ref{fig:india}. A comprehensive result with comparisons with existing state-of-the-art methods were shown in Table 2, this analysis was carried out in KITTI dataset. It can be seen that precision of our lane
detection method at 40m range outperform other existing methods.
\begin{equation}
PRE = \frac{TP}{TP+FP}
\end{equation}
Where TP is true positive and FP is false positive. Also PRE-20, PRE-30 and PRE-40 are precision in 20m, 30m and 40m of IPM image respectively. This precision measure with respect to distance are important are important for lane detection efficiency and depends on IPM image computation.
For lane departure warning system, optical flow computation result are shown in Fig.~\ref{fig:optical}. Optical flow near to vehicles is used for estimating its possible horizontal velocity.

\section{Conclusion}
In this paper, a robust lane detection system is presented using 
steerable edge features and RANSAC polynomial fitting. We have got 
considerable accuracy for lane detection and warning system even 
in shadow and sunny road. This algorithm especially focus on 
enhancing safety in normal driving and for autonomous vehicles
by keeping track of its proper lane, also addresses the problem of 
Indian road condition by defining new boundary detection method. 
In future, we will implement a probabilistic lane tracking system 
for reducing per frame processing cost.
\end{document}